\documentclass[11pt,a4paper]{article}

\usepackage[a4paper,margin=25mm]{geometry}
\usepackage{lmodern}
\usepackage{microtype}
\usepackage{amsmath,amssymb,mathtools,amsthm}
\usepackage{graphicx}
\usepackage{booktabs,tabularx,array}
\usepackage{enumitem}
\usepackage{xcolor}
\usepackage{hyperref}
\usepackage{natbib}
\usepackage{indentfirst}
\usepackage{nicefrac}
\usepackage{float}
\usepackage{comment}
\usepackage[page]{appendix}
\usepackage[nameinlink,noabbrev]{cleveref}

\hypersetup{
    colorlinks=true,
    linkcolor=blue!50!black,
    citecolor=blue!50!black,
    urlcolor=blue!50!black
}

\setlist{nosep,leftmargin=1.6em}
\newcommand{\phy}{\mathrm{phy}}

\title{
\textbf{SocialDreamer: Imagining Social Futures for Individual Decision-Making}
}

\author{
  \textbf{
  Kailai Sun\textsuperscript{1,2,\textdagger}, 
  Xiaochen Chang\textsuperscript{3,\textdagger},
  Ziyou Zhang \textsuperscript{4},
  Baotong Mi\textsuperscript{5}, 
  Yunpeng Chai\textsuperscript{6}
  } \\[4pt]
  \normalfont
  \textsuperscript{1}MIT \quad
  \textsuperscript{2}SMART \quad
  \textsuperscript{3}UCASS\quad
  \textsuperscript{4}THU\quad
  \textsuperscript{5}Renlab AI \quad
  \textsuperscript{6}RUC \\[2pt]
  {\textsuperscript{\textdagger}Equal contribution.}
}

  \normalfont

\date{}
\begin{document}

\maketitle


\begin{abstract}

World models use imagined futures to guide action. However, in social environments, the same proposal can elicit different responses because participants have observed different events or interpret them in light of different beliefs and goals. We introduce SocialDreamer, an ego-centric world model that treats other participants' responses as both drivers of social change and evidence about hidden social states. From its interaction history, the ego agent maintains a distribution over structured state hypotheses that distinguish each participant's physical circumstances, available information, and latent mental state. For each proposed action, SocialDreamer explicitly predicts other participants' possible responses, then the possible successor states after each interaction. The ego agent uses these alternatives and their expected values to refine its action. After execution, it reweights the state hypotheses by comparing its predictions with the observed interactions. Our experiment of this framework uses fixed language models. On SOTOPIA-Hard, SocialDreamer achieves a goal-achievement score of 4.49, outperforming direct inference (3.44) by $30.5\%$ and Foresee and Act (3.58) by $25.4\%$.
\end{abstract}


\section{Introduction}

An agent operating in a social environment must anticipate not only the direct consequences of its own actions, but also how other participants perceive, interpret, and respond to those actions. The same message, gesture, or proposal may lead to different outcomes depending on the recipients' knowledge, beliefs, goals, intentions, emotions, and relationships. Consequently, effective social decision-making requires an agent to reason about the hidden internal states of other participants and anticipate the behavioral responses arising from them. Unlike physical control, where the environment primarily evolves in response to the agent's actions and underlying physical dynamics, social environments involve multiple autonomous participants whose actions jointly shape how the world evolves. Understanding and predicting these interactions is therefore fundamental to effective social decision-making.

World models provide a general approach to decision-making through learned predictions of future states. Dreamer and DreamerV3 demonstrates how an agent can learn compact latent dynamics from partial observations, imagine possible future trajectories, and improve its behavior through latent imagination (Hafner et al., 2020, 2025). Its representation model infers the current latent state from the previous state, preceding action, and newly received observation, while its transition model predicts future states without requiring future observations. This paradigm offers a natural foundation for social decision-making, where an agent could imagine how different actions might influence other participants and the resulting social world. However, extending latent imagination to social environments requires accounting for the distinct perspectives and behaviors of multiple autonomous participants.

In social environments, partial observability is not limited to the ego agent. Every participant has its own physical context, internal mental state, and observation history, and can only act upon the information available to it. A world model may need to maintain information about the hidden states of all participants to predict how the social world evolves, but this does not mean that any individual participant can directly access the complete global state. For instance, one participant may know that an object has been moved, while another remains unaware and continues to believe that it is in its original location. These differences in knowledge can lead to different behaviors even under the same physical circumstances. If all participants' states are encoded in an undifferentiated latent representation and directly exposed to every participant's response model, the resulting predictions may implicitly grant individuals access to information unavailable to them in real interactions. We must therefore preserve not only the physical and mental states of individual participants, but also their distinct information boundaries. 

Social world models make these perspectives explicit. The Foresee and Act procedure built on S$^{3}$AP predicts one possible next social state for a proposed action and uses it to refine that action (Zhou et al., 2026). Mental World Modeling couples physical and mental transitions and evaluates candidate actions through their predicted consequences (Fei and Zhao, 2026). However, the current social state is a partial estimate with several possibilities. Even for a single proposal, several responses may be plausible, each with different consequences for the ego's goal. Keeping track of which state hypotheses predict each response allows the ego to weigh those possibilities before acting and to revisit its hypotheses once it observes what actually happens.

Moreover, the evolution of a social world is fundamentally mediated by interactions among participants. An action initiated by the ego agent does not determine the subsequent social state on its own. Instead, it may change what other participants perceive and believe, eliciting responses that depend on their respective internal states and accessible information. These responses, together with the initiating action, jointly shape the subsequent physical and mental states of the social world. Other participants' actions are therefore not merely external disturbances, but essential components of social state transitions. Importantly, these interactions also provide valuable evidence about the hidden states that generate them. Different mental-state hypotheses may predict different responses to the same action, allowing observed interactions to distinguish between possible underlying social states. Thus, social interactions play a dual role: they drive the evolution of the social world while providing evidence for inferring its latent states.

We present \textbf{SocialDreamer}, an ego-centric decision-making framework that extends world-model-based latent imagination to social environments through structured social representation and agent-conditioned interaction dynamics. SocialDreamer adopts a structured social state that organizes the physical and latent mental information of individual participants:

\[
s_t =\left( s_t^{w,phy}, \{ s_t^{i,phy} \}_{i\in I},\{s_t^{i,ment}\}_{i\in I}\right).
\]
For each participant \(i\), \(s_t^{i,phy}\) represents its physical state and accessible information, while \(s_t^{i,ment}\) captures its latent mental state, including beliefs, goals, intentions, and other behavior-relevant internal factors. The shared physical environment is thus represented in $s_t^{w,phy}$. This structure preserves the ownership of each participant's internal state, while the corresponding information-accessibility constraints ensure that individual mental-state updates and behavioral responses depend only on information available to that participant. The resulting structured representation provides a common state space for social-state estimation and interaction prediction.

The central component of SocialDreamer is the \textbf{Social Predictor}, which explicitly models how interactions generate changes in the social world. Given a social world state and a candidate action from the ego agent, the Social Predictor first constructs an intermediate social state reflecting the action's immediate physical consequences and the corresponding mental-state updates of participants. Other participants then generate responses conditioned on their respective internal states and accessible information. These responses jointly determine the subsequent physical state and the corresponding mental-state updates, yielding a complete physical–mental successor state. By modeling other participants' actions as endogenous responses rather than implicit environmental noise, the Social Predictor generates a joint distribution of state transitions over possible interactions and results. This enables the ego agent to imagine how its actions may unfold through the behaviors of other autonomous participants.

The \textbf{Social Representor} estimates the input latent social state of Social Predictor through the ego's history. Since their internal states are not directly observable, multiple social-state hypotheses may be compatible with the same interaction history. The Representor therefore maintains a probability distribution over possible structured latent states. Each state preserves relevant information to reconstruct observed interactions and predict social dynamics. When a new interaction is observed, the Representor evaluates the compatibility between the new interaction and those predictions under different state hypotheses. Through Bayesian updating, it revises the probabilities of these hypotheses and propagates the resulting distribution to the next social state. This provides an empirical basis for state transitions across rounds, thereby reduces unsubstantiated estimation drift and improves subsequent action selections. In this way, the Social Predictor not only generates possible social futures but also provides the predictive basis for state estimation.

Finally, the \textbf{Social Planner} evaluates the imagined social futures generated by the Social Predictor and refine the action according to the ego agent's objectives. Specifically, the Social Planner constructs candidate actions, and uses the predicted futures to evaluate the expected improvement toward the goal—accounting for uncertainty in both the estimated current social state and possible future interactions—and refine the action. Prediction confidence and behavioral support constraints further account for uncertainty in the learned social dynamics and the limitations of available interaction data. These components enable the ego-centric social decision making through structured state estimation, interaction prediction, and imagined social futures.

Our main contributions are summarized as follows:

1. We formulate a structured social state representation that distinguishes agent-specific physical and latent mental states while preserving individual information boundaries.

2. We develop a Social Predictor and a Social Representor based on interaction. The Social Predictor explicitly models other participants' responses, while the Social Representor uses the compatibility between predicted and observed interactions for Bayesian state estimation updates.

3. We integrated the social prediction and state estimation into a decision making framework that evaluates possible social futures under uncertainty.

\begin{figure}[t]
    \centering
    \includegraphics[width=\textwidth]{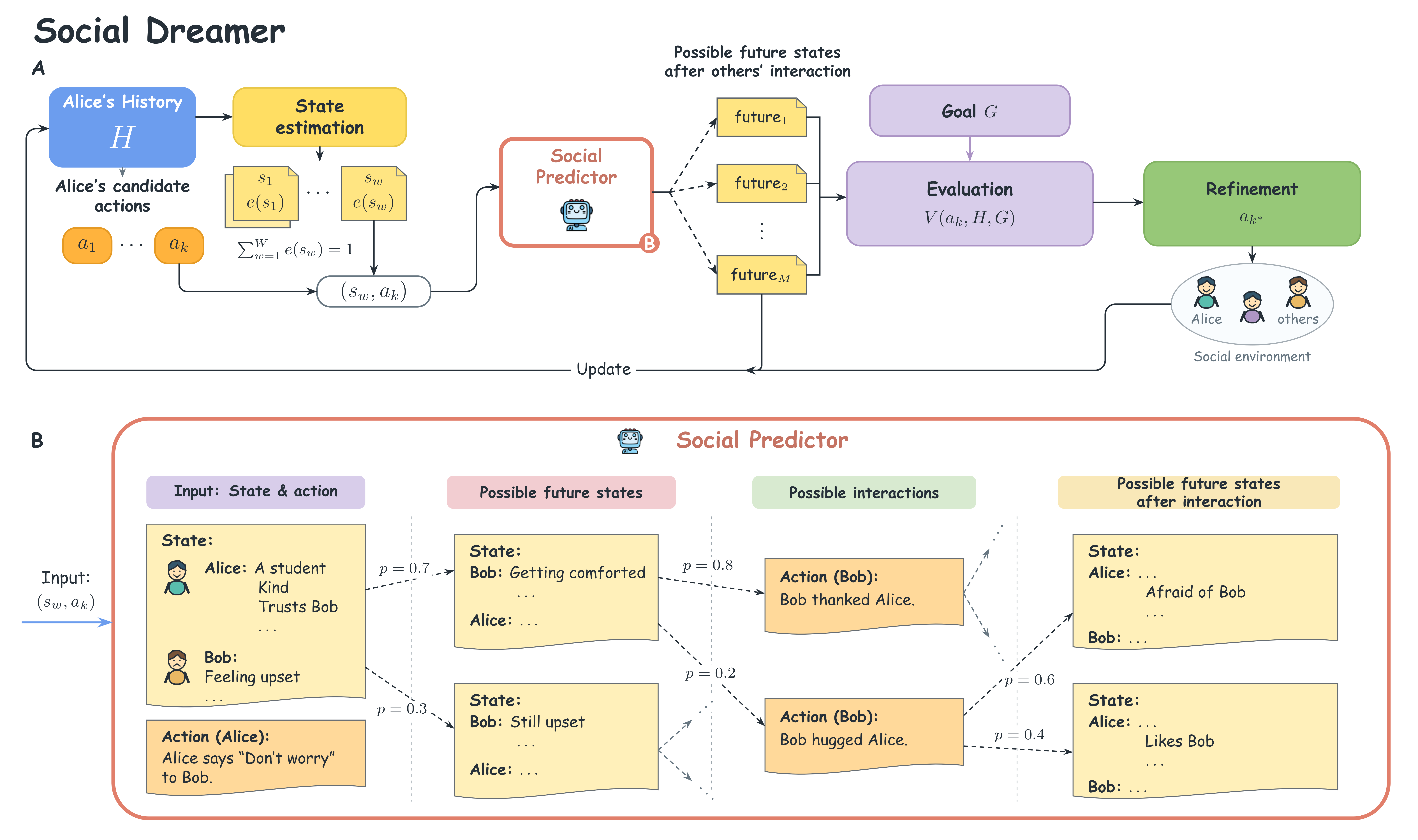}
    \caption{
    \textbf{Overview of SocialDreamer.}
    \text{(A)} Ego-centric social decision making. A distribution over possible latent social states is estimated from the ego agent's interaction history. For each state hypothesis and candidate action, the Social Predictor generates possible social futures, which are then evaluated according to the goal. The ego's action is then refined by the expected return across state hypotheses and predicted outcomes. Observed interactions are subsequently used to update the state estimate. (B) The structure of Social Predictor. Given a social state and an initiating action, the Social Predictor first generates possible intermediate physical–mental states, then predicts other participants' responses conditioned on their internal states and accessible information, and finally generates the resulting social successor states.
    }
    \label{fig:socialdreamer}
\end{figure}

The overall architecture is illustrated in
\cref{fig:socialdreamer}.

\section{Social State Modeling}
\label{sec:state}

This section defines the state space and structural assumptions in SocialDreamer. The objective is to represent a social world containing multiple autonomous participants, each with its own physical and mental state.

\subsection{Structured Social State}

Let $I$ be a finite set of participating agents. We denote the ego agent by $\epsilon\in I$.

We adopt a structured latent representation of the global social state at time $t$:

\begin{equation}
s_t =\left(s_t^{w,phy}, \{ s_t^{i,phy} \}_{i\in I},\{s_t^{i,ment}\}_{i\in I}\right).
\label{eq:global-state}
\end{equation}

The component $s_t^{i,phy}$ contains the physical context available to participant $i$, representing its physical conditions, externally defined attributes and relations, and accessible external information. A physical state of an agent may includes its position, identity, things accessable, communication access, and also its physical relations with others (for instance, its classmates). $s_t^{w,phy}$ then contains other relevant physical conditions in the world (e.g. position of an item). The component $s_t^{i,ment}$ then contains the agent's belief, memory, emotions, mental relations (e.g. trust another agent), goal etc.

The structured factorization preserves the distinction between physical and mental information and the ownership of each participant's internal representation. For each participant $ i \in I$, its own physical and mental latent components is denoted as:

\begin{equation}
s_t^i=\left(s_t^{i,phy},s_t^{i,ment}\right).
\label{eq:agent-state}
\end{equation}

The above factorization specifies the semantic organization of the social state. Each component can be further decomposed into task-dependent attributes. In an explicit representation, these attributes can be organized as structured fields, such as a JSON representation, with their values described in natural language. Alternatively, in a learned representation, the fine-grained information within each component can be encoded into latent variables and learned from interaction data. These alternatives share the same high-level state structure while differing in their representation and learning mechanisms.

The observation $O_t^i$ of each agent $i$ can be reconstructed from $(s_t^i,s_t^{w,phy})$, while its interpretation is determined by $(O_t^i,s_t^{i,ment})$. Thus, the observation represents an agent-specific view of the physical environment, while its interpretation is captured by the agent's mental state. The structured representation specifies the information contained in each component. These components may be instantiated as learned latent variables or explicit structured representations.

\subsection{Actions and Interactions}

An action is an externally manifested physical behavior:

\begin{equation}
a_t^i \in \mathcal A^{\mathrm{phy}}.
\label{eq:action}
\end{equation}

The action space $\mathcal A^{\mathrm{phy}}$ includes utterance content, tone, facial expressions, gestures, movements, timing, and other observable aspects of behavior relevant to the interaction. An action represents the physical and communicative manifestations of behavior, rather than the initiator's unexpressed beliefs, goals, or intentions. Its intended meaning is conveyed through observable signals and physical consequences, which other participants perceive and interpret according to their perceptual circumstances and internal mental states. An action need not be observed by every participant, and a recipient's interpretation need not coincide with the initiator's intention.

We consider an interaction round initiated by the ego agent. The initiating action is denoted by $a_t^\epsilon$, and the other participants' joint response is denoted by

\[
a_t^{-\epsilon}
=
(a_t^i)_{i\in I\setminus\{\epsilon\}}.
\]

For a one-round interaction, the complete action record for the round is

\begin{equation}
a_t = (a_t^\epsilon, a_t^{-\epsilon}).
\label{eq:joint-action}
\end{equation}

This tuple records the initiating action and subsequent responses within the same interaction round. It does not imply that all actions occur simultaneously. Together, these actions constitute the interaction through which participants influence the physical and mental states of the social world.

\subsection{Social State Transitions}

A social state transition describes how the joint social state
changes over one interaction round. Starting from $s_t$, the ego
agent performs an initiating action $a_t^\epsilon$, other
participants respond with $a_t^{-\epsilon}$, and the interaction
results in a successor state $s_{t+1}$. For convenience, we group
the state components as
\[ s_t^{phy} = \left(s_t^{w,phy},\{s_t^{i,phy}\}_{i\in I}\right),
\qquad
s_t^{ment}=\{s_t^{i,ment}\}_{i\in I}.\]
So that $s_t=(s_t^{phy},s_t^{ment})$.

Let $P_\theta$ denote the predictive kernel of social state transition, which models the distribution of successor states following an ego action. The symbol $\theta$ specifies the predictive model, including learnable parameters in a learned model, or the associated prompt and decoding configuration in a pretrained fixed model. The successor state is denoted as:
\begin{equation}
s_{t+1} \sim P_\theta(\cdot\mid s_t,a_t^\epsilon).
\end{equation}
Here, the responses of other participants are marginalized, rather than supplied as actions controlled by the ego agent.

The physical and mental successors are not always independent. Under the same current state and ego action, different physical outcomes may be associated with different subsequent beliefs, emotions, or intentions. Providing both successors with the same current inputs does not always preserve this dependence. The mental successor is conditioned on the physical successor of the same predictive branch.\footnote{Consider an idealized setting in which the ego agent A presses a door-opening button and the door opens with probability $1/2$. Another agent B accurately observes the outcome and forms the belief that the door is either open or closed. The belief of agent B should correspond to the outcome in that branch. However, independently sampling the physical and mental successor does not satisfy this result. Conditioning the mental successor on the corresponding physical successor preserves this dependence.} We therefore represent the joint distribution through conditional generation:

\begin{equation}
\begin{aligned}
P_\theta (s_{t+1}\mid s_t,a_t^\epsilon) = &P_\theta^{phy} (s_{t+1}^{phy}\mid s_t,a_t^\epsilon)\\
&\times P_\theta^{ment} (s_{t+1}^{ment} \mid s_{t+1}^{phy},s_t,a_t^\epsilon).
\end{aligned}
\label{eq:joint-transition}
\end{equation}

Equation \eqref{eq:joint-transition} is the chain rule factorization of the same joint distribution. It allows the dependence between physical and mental successors to be represented within the same predictive branch. The ordering does not establish a causal direction.

When referring to a model-generated future state, we use a hat to identify a predicted state:
\[
\hat s_{t+1} = \left( \hat s_{t+1}^{phy}, \hat s_{t+1}^{ment}\right).
\qquad \hat s_{t+1} \sim P_\theta(\cdot\mid s_t,a_t^\epsilon).
\]

\subsection{Markovian Social State}

We model the complete social state as Markovian at the temporal resolution of one interaction round. The Markov formulation assumes that the current state retains all history-dependent information relevant to prediction. Specifically, conditional on the current global social state and action, the successor state is independent of earlier states and actions:
\begin{equation}
P_\theta(s_{t+1}\mid s_0,\dots,s_{t},a_0,\dots,a_{t-1},a_t^\epsilon)=P_\theta(s_{t+1}\mid s_t,a_t^\epsilon).
\label{eq:markov}
\end{equation}
Here, $a_0,\dots,a_{t-1}$ denote the complete action records of previous interaction rounds, while $a_t^\epsilon$ denotes the action record of the ego agent in this current round. The current responses of other participants are marginalized in the transition kernel $P_\theta(\cdot\mid s_t,a_t^\epsilon)$. Their response mechanisms must therefore depend on earlier interactions only through information retained in the current state.

From the ego agent's perspective, however, the global social state is only partially observable. The ego cannot directly access other participants' internal mental states or hidden information. This setting can be formulated as a partially observable Markov decision process (POMDP), in which the global social state evolves according to the transition kernel $P_\theta$, while the ego receives observations determined by its own physical circumstances and accessible information.

Let $H_t^\epsilon$ denote the interaction history available to the ego before its interaction round $t$. The history contains all information available to the ego, including ego's own personal information, retained experiences, previous interactions, and observations of other participants also the physical environment. The history may contain multimodal records including text, audio, images, videos etc. During a decision round, $H_t^\epsilon$ remains fixed across candidate actions and predictive branches. Historical information used for prediction is encoded in the current state hypothesis $s_t$. The Social Predictor will take only $s_t$ and the candidate ego action $a_t^\epsilon$ as its inputs.

Rather than directly conditioning its decisions on the unknown global state, the ego maintains a belief distribution over possible social states:
\begin{equation}
b_t(s_t) = R_\phi(s_t\mid H_t^\epsilon).
\label{eq:belief-state}
\end{equation}
Here, $H_t^\epsilon$ denotes the complete observation and interaction history available to the ego. The distribution $R_\phi$ therefore serves as the ego's belief state, representing its uncertainty about the global social state. The Social Representor updates this belief when new interaction evidence becomes available, as detailed in Section~\ref{sec:representor}.

This assumption requires $s_t$ to preserve historical information relevant to future social dynamics. Such information may include participants' memories, acquired beliefs, goals, commitments, persistent relationships, relevant knowledge acquired through earlier interactions, and changes in their behavioral tendencies. The state could therefore retain the information needed to predict subsequent responses and state transitions.

Markovianity is a modeling assumption about the predictive sufficiency of the state representation. The state is intended to retain the historical
information relevant to future social dynamics. Unlike conventional POMDP formulations that treat other participants' behaviors as part of the environment dynamics (Kaelbling et al., 1998), SocialDreamer explicitly models their agent-specific physical and mental states and retains their responses as intermediate predictive variables. This enables interaction outcomes to serve both as
state transitions and as evidence for belief updating.

\section{Social Predictor}
\label{sec:predictor}

The Social Predictor is the central component of SocialDreamer. It models the transition kernel introduced in Section 2 through a two-stage interaction process. Given a current-state hypothesis $s_t$ and a candidate ego action $a_t^\epsilon$, the Social Predictor outputs the prediction of other agents' responses $\hat a_t^{-\epsilon}$ and the prediction of the successor state $\hat s_{t+1}$. Specifically, it first predicts the direct result of the initiating action into an intermediate physical-mental state, including how other participants may observe and interpret the result and update their mental states. Then it predicts the other participants' responses and the resulting successor state. The joint distribution supports social imagination and provides the predictive basis for the state estimation.

The one-round prediction of the Social Predictor follows the structure:
\begin{equation}
s_t \xrightarrow{a_t^\epsilon} \hat s_{t+\nicefrac12}
\xrightarrow{\hat a_t^{-\epsilon}} \hat s_{t+1}.
\label{eq:two-stage}
\end{equation}
Index $t+\nicefrac12$ denotes a logical intermediate stage within one interaction round. The predicted round terminates at the complete successor state $\hat s_{t+1}$.

\subsection{Initial Action and Intermediate State}

The first stage predicts the intermediate social state following the ego's initiating action, before the other participants generate their responses. We denote the first-stage transition kernel by $P_\theta^{(1)}$.

The initiating action will first results a direct physical update:

\begin{equation}
\hat s_{t+\nicefrac12}^{phy} \sim P_\theta^{(1),phy}
(\cdot\mid s_t^{phy},a_t^\epsilon).
\label{eq:stage1-physical}
\end{equation}
This prediction represents the immediate physical and communicative consequences of the initiating action, including changes in the external information available to individual participants. Conditional on the physical state and the fully specified externally manifested action, private mental states have no additional direct influence on the immediate physical update. 

Conditioned on the physical outcome in the same predictive branch, we then predict the intermediate mental state:
\begin{equation}
\hat s_{t+\nicefrac12}^{ment} \sim P_\theta^{(1),ment}
\left( \cdot\mid s_t,a_t^\epsilon,\hat s_{t+\nicefrac12}^{phy}\right).
\label{eq:stage1-mental}
\end{equation}
For each participant, its mental update should respect its previous internal state and the information available to it within that branch.

The intermediate social state is
\[
\hat s_{t+\nicefrac12} = \left(\hat s_{t+\nicefrac12}^{phy},\hat s_{t+\nicefrac12}^{ment}\right).
\]

This intermediate state provides the context for predicting the other participants' responses in the second stage. The corresponding joint distribution in the first stage is:

\begin{equation}
\begin{aligned}
P_\theta^{(1)}
\left( \hat s_{t+\nicefrac12}\mid s_t,a_t^\epsilon \right)
& = P_\theta^{(1),phy} \left(\hat s_{t+\nicefrac12}^{phy} \mid s_t^{phy},a_t^\epsilon\right)\\
&\times P_\theta^{(1),ment} \left( \hat s_{t+\nicefrac12}^{ment}
\mid s_t,a_t^\epsilon,\hat s_{t+\nicefrac12}^{phy} \right).
\end{aligned}
\label{eq:stage1-transition}
\end{equation}

\subsection{Participant Responses and Successor State}

The second stage predicts how the other participants respond to the initiating action and how their responses change the social state. Given the intermediate state generated in the first stage, we predict the joint response as
\begin{equation}
    \hat a_t^{-\epsilon} \sim \pi_\theta^{-\epsilon} \left( \cdot\mid
    s_t,a_t^\epsilon,\hat s_{t+\nicefrac12} \right).
\label{eq:stage2-responses}
\end{equation}
In which $\pi_\theta^{-\epsilon}$ denotes the predictive distribution over the other participants' joint response. Each participant's response is conditioned only on its own internal state and the information available to it within the predicted branch.

Given the predicted responses, we generate the physical
successor:
\begin{equation}
\hat s_{t+1}^{phy} \sim P_\theta^{(2),phy} \left( 
    \cdot\mid \hat s_{t+\nicefrac12}^{phy},\hat a_t^{-\epsilon} \right).
\label{eq:stage2-physical}
\end{equation}

While the mental successor is generated as:
\begin{equation}
\hat s_{t+1}^{ment} \sim P_\theta^{(2),ment} \left( \cdot\mid 
\hat s_{t+\nicefrac12}, \hat a_t^{-\epsilon}, \hat s_{t+1}^{phy} \right).
\label{eq:stage2-mental}
\end{equation}

The complete successor state is
\[
\hat s_{t+1} = \left( \hat s_{t+1}^{phy}, \hat s_{t+1}^{ment} \right).
\]

The corresponding conditional joint distribution of
physical and mental successors is

\begin{equation}
\begin{aligned}
P_\theta^{(2)} \left(\hat s_{t+1} \mid \hat s_{t+\nicefrac12},
\hat a_t^{-\epsilon} \right)
& = P_\theta^{(2),phy} \left(\hat s_{t+1}^{phy} \mid \hat s_{t+\nicefrac12}^{phy},\hat a_t^{-\epsilon} \right)\\
&\times P_\theta^{(2),ment} \left(\hat s_{t+1}^{ment}\mid \hat s_{t+\nicefrac12}, \hat a_t^{-\epsilon}, \hat s_{t+1}^{phy} \right).
\end{aligned}
\label{eq:stage2-transition}
\end{equation}

Here, the intermediate state and the predicted joint response are assumed to retain the information needed to predict the successor state. 

\subsection{Joint Transition Distribution and Marginalization}

The two-stage process induces a joint predictive distribution over the other participants' responses and the resulting successor state. Given a current-state hypothesis $s_t$ and an initiating action $a_t^\epsilon$, we denote this distribution by $K_\theta$:

\begin{equation}
\left( \hat a_t^{-\epsilon},\hat s_{t+1} \right) \sim
K_\theta \left( \cdot,\cdot\mid s_t,a_t^\epsilon \right).
\label{eq:joint-predictive-kernel}
\end{equation}
The kernel $K_\theta$ retains the association between each possible response and its corresponding successor states, which is an intermediate probabilistic description constructed from $P_\theta^{(1)}$, $\pi_\theta^{-\epsilon}$, and
$P_\theta^{(2)}$. Let $\mathcal S$ denote the structured social state space.
Marginalizing the intermediate state, we have:
\begin{equation}
\begin{aligned}
K_\theta \left( \hat a_t^{-\epsilon},\hat s_{t+1} \mid s_t,a_t^\epsilon \right)
& = \int_{\mathcal S} P_\theta^{(1)} \left(\hat s_{t+\nicefrac12} \mid s_t,a_t^\epsilon \right)\\
&\times \pi_\theta^{-\epsilon} \left(\hat a_t^{-\epsilon} \mid s_t,a_t^\epsilon,\hat s_{t+\nicefrac12} \right)\\
&\times P_\theta^{(2)} \left( \hat s_{t+1} \mid \hat s_{t+\nicefrac12},\hat a_t^{-\epsilon} \right) \mathrm d\hat s_{t+\nicefrac12}.
\end{aligned}
\label{eq:joint-predictive-distribution}
\end{equation}

The one-round state transition kernel is then obtained by further marginalizing the other participants' responses:

\begin{equation}
P_\theta \left( \hat s_{t+1}\mid s_t,a_t^\epsilon\right) =\int_{\mathcal A^{-\epsilon}}K_\theta\left(\hat a_t^{-\epsilon},\hat s_{t+1}\mid s_t,a_t^\epsilon
\right)\mathrm d\hat a_t^{-\epsilon}.
\label{eq:statetransition-marginal}
\end{equation}
In which $\mathcal A^{-\epsilon}$ denotes the joint response space of the other participants.

Conversely, marginalizing the successor state yields the response distribution conditioned only on the initial state and ego action:

\begin{equation}
\pi_{\theta,\mathrm{marg}}^{-\epsilon} \left(\hat a_t^{-\epsilon}\mid s_t,a_t^\epsilon\right) =\int_{\mathcal S}K_\theta\left(\hat a_t^{-\epsilon},\hat s_{t+1}\mid s_t,a_t^\epsilon\right)\mathrm d\hat s_{t+1}.
\label{eq:response-marginal}
\end{equation}

The joint kernel $K_\theta$ retains which responses are associated with which successor states. The response marginal $\pi_{\theta,\mathrm{marg}}^{-\epsilon}$ providing the predictive basis for updating state estimates from observed interactions in Section~\ref{sec:representor}. The state marginal $P_\theta$ describes possible social futures for decision-making in Section~\ref{sec:planner}. The expressions above use probability-mass or density notation. Integrals represent marginalization and are replaced by sums for discrete variables.

\subsection{Finite Representation of Social Futures}

Exact marginalization over all possible intermediate states,
participant responses, and successor states may be
computationally intractable. We therefore employ Monte Carlo
sampling to approximate expectations under the predictive
distributions.

Given a current-state hypothesis $s_t$ and a candidate
ego action $a_t^\epsilon$, we generate $M$ independent
predictive branches by sequentially sampling:

\begin{equation}
\begin{aligned}
\hat s_{t+\nicefrac12}^{(m)}&\sim P_\theta^{(1)}(\cdot\mid s_t,a_t^\epsilon),\\
\hat a_t^{-\epsilon,(m)}&\sim \pi_\theta^{-\epsilon}(\cdot\mid s_t,a_t^\epsilon,
\hat s_{t+\nicefrac12}^{(m)}),\\
\hat s_{t+1}^{(m)}&\sim P_\theta^{(2)}(\cdot\mid\hat s_{t+\nicefrac12}^{(m)},
\hat a_t^{-\epsilon,(m)}),
\end{aligned}
\label{eq:mc-sampling}
\end{equation}
for $m=1,\ldots,M$. Each response-successor pair $(\hat a_t^{-\epsilon,(m)},\hat s_{t+1}^{(m)})$ is sampled from the joint kernel $K_\theta$. 

The sampled branches preserve the correspondence between participant responses and their resulting social states. These samples can subsequently be used for expected evaluation or state estimation. For any integrable evaluation or estimation function $f$, its expectation under the joint predictive distribution can be approximated by

\begin{equation}
\mathbb{E}_{K_\theta}[f] \approx \frac{1}{M}\sum_{m=1}^{M} f(\hat a_t^{-\epsilon,(m)},\hat s_{t+1}^{(m)}).
\label{eq:mc-expectation}
\end{equation}

Alternatively, when the predictive components are instantiated using fixed pretrained language models, the possible futures can be represented directly as a finite set of weighted branches. For a current-state hypothesis $s_t$ and a candidate ego action $a_t^\epsilon$, let $\mathcal B_t(a_t^\epsilon)$ denote the generated set of possible futures. For each discrete branch $b$, its probability $p_b$ is obtained by combining the conditional probabilities along the corresponding generation path:
\begin{equation}
p_b={}P_\theta^{(1)}\left(\hat{s}_{t+\nicefrac12,b}\mid s_t,a_t^\epsilon
\right)\times\pi_\theta^{-\epsilon}\left(\hat{a}_{t,b}^{-\epsilon}\mid s_t,a_t^\epsilon,\hat{s}_{t+\nicefrac12,b}\right)\times P_\theta^{(2)}\left(\hat{s}_{t+1,b}\mid\hat{s}_{t+\nicefrac12,b},\hat{a}_{t,b}^{-\epsilon}\right).
\label{eq:explicit-branch-probability}
\end{equation}

The probabilities of the generated branches are then normalized over the finite set, denoted as $\tilde p_b$. The expectation of an evaluation function can be approximated by the probability-weighted sum:
\begin{equation}
\mathbb E[f]\approx\sum_{b\in\mathcal B_t(a_t^\epsilon)}
\tilde p_b f\left(\hat{a}_{t,b}^{-\epsilon},\hat{s}_{t+1,b}\right).
\label{eq:explicit-branch-expectation}
\end{equation}

The two approaches provide alternative finite approximations of the predictive distribution. The latter is adopted in the fixed-model implementation of this study.

\section{Social Representor}
\label{sec:representor}

The Social Predictor generates possible interactions and successor states from a given current-state hypothesis. The Social Representor supplies this hypothesis by maintaining a distribution over structured latent social states inferred from the ego's history. This latent representation is intended to retain the information needed to predict future social dynamics, while the distribution represents the ego's uncertainty about the current social state. The Representor also update the distributions through multi-round interactions. When receiving newly observed interaction evidence, it combines the previous state estimate with the Social Predictor's distribution to update the estimate through Bayesian inference. This provides an empirical basis for state transitions across rounds, thereby reduces unsubstantiated estimation drift and improves subsequent action selections.

\subsection{History, State Estimation and Initialization}

Let $H_0^\epsilon$ denote the initial history of the current estimation process, it contains the information acquired before the interaction rounds considered here.

We describe the Social Representor by a state-estimation kernel $R_\phi$, where $\phi$ specifies its implementation. Given $H_t^\epsilon$, the kernel defines a distribution $R_\phi(\cdot\mid H_t^\epsilon)$ over the structured
social state space $\mathcal S$. Current state hypotheses can be drawn from this distribution:

\begin{equation}
s_t \sim R_\phi(\cdot\mid H_t^\epsilon).
\label{eq:state-estimation-kernel}
\end{equation}
Each hypothesis $s_t$ has the structure defined in Equation~\eqref{eq:global-state}, comprising the shared physical state and the physical and mental states of individual participants. Multiple hypotheses may be compatible with the same available history. The distribution represents uncertainty
over these hypotheses and can retain dependencies among their state components. In particular, other participants' internal states are inferred from the evidence available to the ego.

At initialization, the Representor constructs $R_\phi(\cdot\mid H_0^\epsilon)$ from the initial history $H_0^\epsilon$.  A state hypothesis drawn from this distribution, together with a candidate ego action $a_0^\epsilon$, provides the input to the Social Predictor. The same interface applies at subsequent rounds using the updated distribution $R_\phi(\cdot\mid H_t^\epsilon)$.

After the interaction round $a_t$, let $\Delta H_{t+1}^\epsilon$ denote the newly available record, including new observations and interactions available. The history is extended as:
\begin{equation}
H_{t+1}^\epsilon = H_t^\epsilon\oplus\Delta H_{t+1}^\epsilon,
\end{equation}

When new interaction evidence becomes available, the Representor uses the Social Predictor's response distribution and joint response-successor kernel to update this estimate, while the complete history remains available to the Representor as context. The next subsection defines this Bayesian update.

\subsection{Bayesian State Update with New Interactions}
\label{sec:bayesian-update}

During a multi-round interactions, at the beginning of round $t$, the Representor generates the state estimate distribution $R_\phi(\cdot\mid H_t^\epsilon)$. For each current state hypothesis $s_t$ and the ego's initiating action $a_t^\epsilon$, the Social Predictor provides the distribution of other participants' responses through $\pi_{\theta,\mathrm{marg}}^{-\epsilon} (\cdot\mid s_t,a_t^\epsilon)$ defined in Equation~\eqref{eq:response-marginal}, and also the joint kernel $K_\theta$ in \eqref{eq:joint-predictive-kernel}.

A newly observed interaction \(\Delta H_{t+1}^\epsilon\) provides evidence about the latent state from which it arose. The Representor first uses this evidence to revise its estimate of the preceding state $s_t$, and then propagates the resulting posterior to estimate the successor state $s_{t+1}$. The learned model parameters remain fixed during this recursive update.

\begin{figure}[H]
    \centering
    \includegraphics[width=0.65\textwidth]{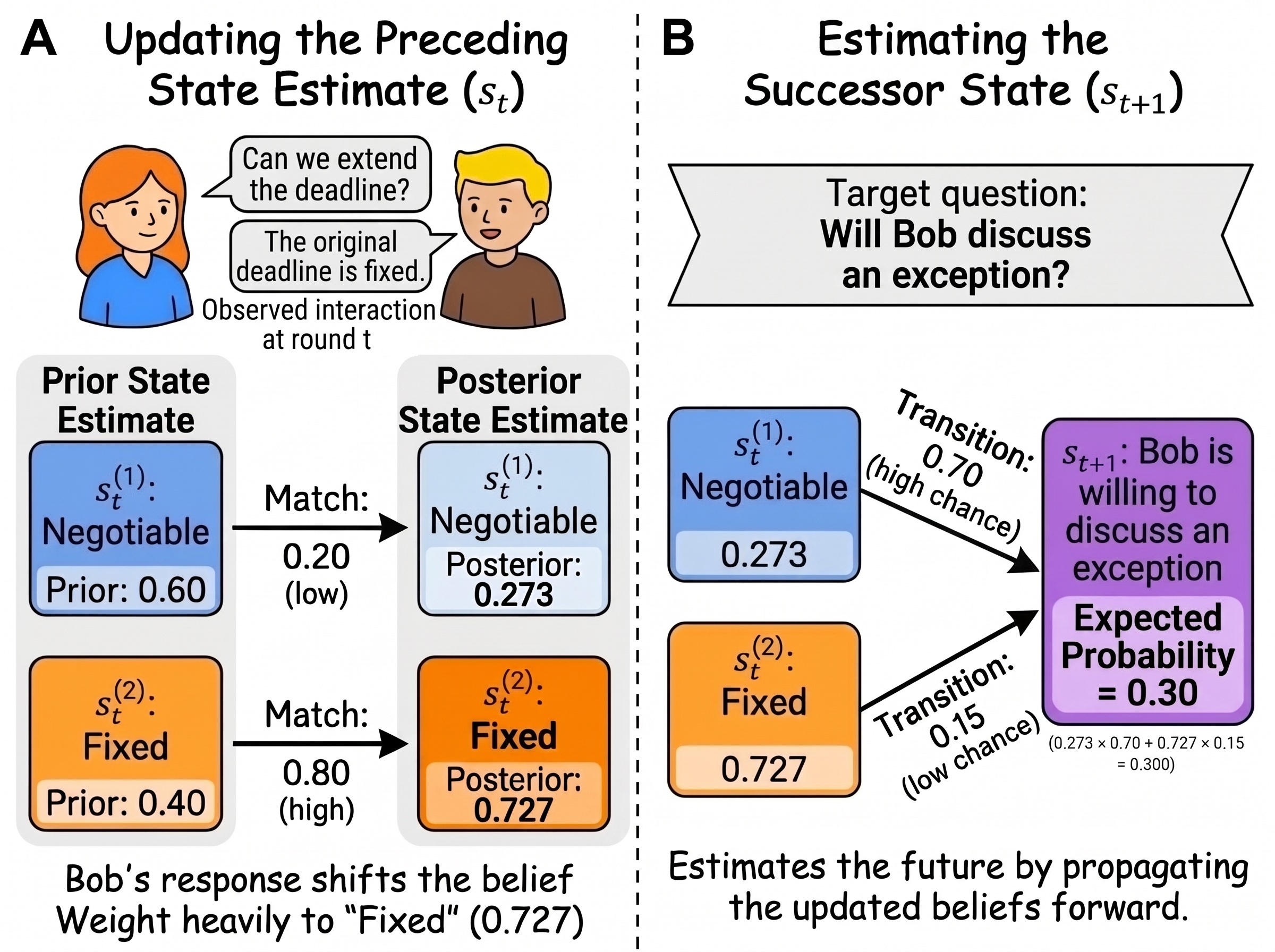}
    \caption{
    \textbf{Bayesian state estimation with new interactions.}
    Alice asks Bob to extend a deadline. (A) Bob's reply provides evidence about his preceding mental state, revising the probabilities of state hypotheses. (B) The revised weights are combined with successor probabilities conditioned on the same reply to estimate the  probability that Bob is willing to discuss an exception.
    }
    \label{fig:bayesian-state-update}
\end{figure}

Figure~\ref{fig:bayesian-state-update} illustrates the bayesian update for a single observed interaction. Panel A revises the probabilities of the preceding-state hypotheses, and Panel B uses the revised weights and the response-conditioned successor branches to estimate the successor state.

Under a possible branch, We evaluate the likehood of the observed record under the predicted interaction, using the matching weight approximated by a matching model:
\begin{equation}
M^\epsilon(\Delta H_{t+1}^\epsilon,\hat a_t) \in (0,1], \quad \hat a_t=(a_t^\epsilon, \hat a_t^{-\epsilon}).
\label{eq:likehood}
\end{equation}
where the ego's action is fixed and the other participants' responses are predicted.

For each state hypothesis, the predicted response probabilities are weighted by their compatibility with the new record and integrated over possible responses. Using Bayes' rule, we obtain the posterior distribution of state estimation from the prior:
\begin{equation}
R_\phi(s_t\mid H_{t+1}^\epsilon) = \frac{ \displaystyle 
R_\phi(s_t\mid H_t^\epsilon) \int_{\mathcal A^{-\epsilon}} \pi_{\theta,\mathrm{marg}}^{-\epsilon} \left( \hat a_t^{-\epsilon}\mid s_t,a_t^\epsilon\right)M^\epsilon\left(\Delta H_{t+1}^\epsilon,\hat a_t\right)\mathrm d\hat a_t^{-\epsilon}}
{\displaystyle\int_{\mathcal S}R_\phi(s_t\mid H_t^\epsilon)\left[\int_{\mathcal A^{-\epsilon}}\pi_{\theta,\mathrm{marg}}^{-\epsilon}\left(\hat a_t^{-\epsilon}\mid s_t,a_t^\epsilon\right)M^\epsilon\left(\Delta H_{t+1}^\epsilon,\hat a_t
\right)\,\mathrm d\hat a_t^{-\epsilon}\right]\mathrm ds_t}.
\label{eq:preceding-state-posterior}
\end{equation}
The numerator combines the prior probability of the state hypothesis, its predicted response probabilities, and the compatibility of those responses with the observed record. The denominator normalizes these weights over the state space. A state hypothesis therefore receives greater posterior support when it assigns higher probability to interactions that better explain the new evidence. The distribution $R_\phi(s_t\mid H_t^\epsilon)$ is the prior relative to the new record, whereas $R_\phi(s_t\mid H_{t+1}^\epsilon)$ is the posterior after that record has been incorporated. Both distributions refer to the same state $s_t$. 

The revised estimate is then propagated to $s_{t+1}$ through the existing predictive model. This propagation retains the constraints imposed by the same interaction evidence. For each preceding-state hypothesis, the joint kernel $K_\theta$ associates possible responses with their successor states. Weighting these predictions by $M^\epsilon$ and normalizing gives the conditional distribution of successor states under the observed record. Averaging over the revised estimate of $s_t$ yields:
\begin{equation}
\begin{aligned}
R_\phi(s_{t+1}\mid H_{t+1}^\epsilon) = & \int_{\mathcal S}R_\phi(s_t\mid H_{t+1}^\epsilon)\\
& \times \frac{\displaystyle \int_{\mathcal A^{-\epsilon}}K_\theta\left(\hat a_t^{-\epsilon},\hat s_{t+1}\mid s_t,a_t^\epsilon\right)M^\epsilon\left(\Delta H_{t+1}^\epsilon,\hat a_t\right)\mathrm d\hat a_t^{-\epsilon}}
{\displaystyle \int_{\mathcal A^{-\epsilon}}\pi_{\theta,\mathrm{marg}}^{-\epsilon}\left(\hat a_t^{-\epsilon}\mid s_t,a_t^\epsilon\right)M^\epsilon
\left(\Delta H_{t+1}^\epsilon,\hat a_t\right)\mathrm d\hat a_t^{-\epsilon}}
\mathrm ds_t.
\end{aligned}
\label{eq:posterior-state-propagation}
\end{equation}

This procedure updates the state estimate. The resulting $R_\phi(\cdot\mid H_{t+1}^\epsilon)$ provides the state estimate for the next interaction round. It provides an empirical basis for state transitions across rounds, thereby reduces unsubstantiated estimation drift and improves subsequent action selections. Integrals are replaced by sums for discrete variables.

\section{Social Planner}
\label{sec:planner}

The Social Planner generates candidate actions and evaluate their anticipated consequences, then refines the proposals into a single action for the ego to execute according to its goal. With the help of the Social Predictor and the Social Representor, we can now evaluate the expected value of each candidate action. The Planner can learn in a trainable implementation, or use a fixed language model to perform action proposal, outcome evaluation, and refinement through structured prompts.

We first define this evaluation and refinement process, then examine the applicability of behavioral prediction to rational planning and the conditions for reliable decisions.

\subsection{Action Proposal and Refinement}
\label{sec: evaluation}

Let $G$ denote the ego's goal. At the beginning of round $t$, the ego evaluates candidate actions $\mathcal A_t^\epsilon=\{a_t^{\epsilon,k}\}$ using the current state estimate $R_\phi(\cdot\mid H_t^\epsilon)$, where $|\mathcal A_t^{\epsilon}| \geq 1$.

With a history $H_t^\epsilon$ and a goal $G$ for the ego agent, for each candidate action $a_t^{\epsilon,k} \in\mathcal A_t^\epsilon$, let $V(a_t^{\epsilon,k},H_t^\epsilon,G)$ denote its predicted value under goal $G$. This scalar value represents the action's expected contribution to goal achievement, including both immediate progress and long-term consequences. The evaluation function is denote as:
\begin{equation}
    V(a_t^{\epsilon,k},H_t^\epsilon,G)=\int_{\mathcal S}R_\phi(s_t\mid H_t^\epsilon)
    \left[\int_{\mathcal S} P_\theta(\hat s_{t+1}\mid s_t,a_t^{\epsilon,k})
    V(s_t,\hat s_{t+1},G) \mathrm d\hat s_{t+1} \right]
\mathrm ds_t.
\label{eq:value}
\end{equation}
In which $V(s_t,\hat s_{t+1},G)$ denotes how state $\hat s_{t+1}$ satisfies the goal than $s_t$, both long-term and short-term. We use $V$ for both transition values and action values, distinguished by their arguments. We will introduce the detailed valuation function in Appendix ~\ref{app:value}.

The expected values provide the basis for refining the initial action proposals. The Social Planner considers the predicted responses and successor states associated with each candidate, together with their corresponding values, to identify potential improvements and undesirable consequences. It then revises the proposed actions to better account for these anticipated social outcomes and improve their expected contribution to the ego's goal.

The final action is produced based on the initial candidates and their predicted futures. When the benchmark provides predefined actions, the ego selects the best action among them through the evaluation. In open-ended settings, the Planner may modify the content of the initial proposals to produce an improved action. If refinement changes the action, the Predictor generates the interaction branches used for subsequent state updating under the final action before it is executed.

\subsection{From Behavior Prediction to Rational Planning}
The Social Predictor models how participants respond to a given initiating action. These behavioral predictions can be learned from human interaction data or elicited from a pretrained language model through structured prompts. It does not require the initiating participant to be a rational optimizer. However, when the ego uses the predictions to refine its action proposals according to its goal, he becomes a rational planner. We examine which predictions remain applicable when an initiating action is held fixed while the ego's private mental state and decision process may differ.

Let $s_t^{-\epsilon,ment}$ denote the mental states of all participants other than the ego. We define the equivalence relation
\begin{equation}
s_t\equiv {s'}_t \iff \bigl( s_t^{phy}={s'}_t^{phy} \wedge s_t^{-\epsilon,ment}={s'}_t^{-\epsilon,ment} \bigr).
\end{equation}
The equivalent states therefore share the same physical conditions and the same mental states of other participants, while allowing the ego's private mental state to differ. Let $[s]$ denote the equivalence class containing $s$.

We assume that the initiating action specifies all relevant externally expressed behavior and that the ego performs no additional physical action within the same interaction round. Thus, the ego's unexpressed private mental state have no additional direct influence on the joint mental updates and responses of other participants. For the transition kernel in \eqref{eq:statetransition-marginal}, we have:

\begin{equation}
    s_t \equiv s_t' \Rightarrow
    \bigl(\forall a_t^\epsilon\in\mathcal A^\phy,
    P_\theta(\cdot \mid s_t,a_t^\epsilon) \equiv P_\theta(\cdot \mid s_t',a_t^\epsilon)\bigr).
\end{equation}

To see this, equivalent initial states and the same initiating action produce the same intermediate physical distribution. The information constraints then preserve the joint distribution of the other participants' mental updates and responses. These responses generate the same distribution of physical successors and other participants' mental successors. The ego's own mental successor may differ, but marginalizing this component gives the same distribution over successor equivalence classes.

We can therefore define an induced transition kernel on $\mathcal S/\equiv$:

\begin{equation}
    \bar P_\theta \left([\hat s_{t+1}]\mid[s_t],a_t^\epsilon\right) =\int P_\theta(\hat s_{t+1}\mid s_t,a_t^\epsilon)\mathrm d\hat s_{t+1}^{\epsilon,ment}.
\end{equation}

This property supports reusing conditional predictions of physical and other participant consequences when the ego adopts a different action-selection procedure. The equivalence result establishes structural compatibility of the predictive mechanisms.

\subsection{Behavioral Support and Predictive Confidence}
\label{sec:planning-bounds}

For a fixed initiating action, the equivalence result in Section~5.2 determines the distribution over successor equivalence classes. Within a given class, the rational ego's private mental successor may differ from that predicted for an ordinary participant. The Planner therefore requires that the rational ego's predicted successor have sufficient conditional probability under the modeled behavior of ordinary participants.

For a current-state hypothesis $s_t$ and a candidate action $a_t^{\epsilon,k}$, let $\hat s_{t+1}$ denote a successor predicted under ordinary participant behavior, and let $\tilde s_{t+1}\equiv\hat s_{t+1}$ denote the corresponding predicted social state when the ego uses the Planner. The two successors share the same physical state and the same mental states of all other participants. For discrete finite states, we impose the following lower bound for the ego's predicted state:
\begin{equation}
    \frac{ P_\theta(\tilde s_{t+1}\mid s_t,a_t^{\epsilon,k}) }{ \bar P_\theta ([\hat s_{t+1}]\mid[s_t],a_t^{\epsilon,k}) } \geq \rho .
\label{eq:mental-support-threshold}
\end{equation}
This ratio represents the conditional predictive support for the ego's mental successor given the physical and other-participant outcomes. The threshold $\rho>0$ specifies the minimum behavioral support required to retain this prediction. In our first experiment, we simplify $\tilde s_{t+1}$ as $\hat s_{t+1}$.

Predictive support is assessed separately from value. The support score for a predicted branch reflects compatibility with independent interaction evidence and testable constraints on physical mechanisms, temporal order, and information availability. The Planner can use this score to determine which futures to consider during action refinement. Appendix~\ref{app:confidence} gives an optional extension that incorporates predictive confidence into action values and derives a Monte Carlo approximation for their evaluation.

\section{Experiments}
\label{sec:experiments}

This section specifies experiments for evaluating the overall framework and its main modeling components. We first evaluate whether structured social representations and explicit predictions of future interactions improve social decision making without model post-training. Specifically, we aim to answer the following research questions:

\begin{enumerate}
    \item Does SocialDreamer help decision making by explicitly expressing the participants' latent mental states, together with explicit future predicting and current state estimate?

    \item During future predicting, does explicitly predicting other participants'
    responses and the resulting interactions improve action selection compared with directly predicting successor states?

    \item Does updating the state estimation with predicted interactions through Bayesian helps improve the goal achievement decision making over multiple rounds?
\end{enumerate}

To evaluate the structural design of SocialDreamer without confounding effects from model post-training, we instantiated the framework using fixed language models with explicit structured states.

\subsection{Experimental Design}
\paragraph{Benchmark.}

We evaluate SocialDreamer on SOTOPIA-Hard, a benchmark of multi-round, open-ended social interactions. Each participant acts according to its assigned character and private goal. At each turn, the ego produces an action based on its available interaction history. SOTOPIA-Hard contains 100 groups of data\footnote{In which five cases are "revenge and hurting others", and another five are "selling item that doesn't own". We only use the rest 90 groups in our experiment.}, each allows multi-round interactions with at most 20 steps.

\paragraph{Baselines.}

We compare SocialDreamer with two baselines:

\begin{enumerate}
    \item Direct inference. The underlying language model receives the ego's character information, private goal, and available interaction history, and directly generates the next utterance or action.

    \item S$^{3}$AP (Zhou et al., 2026). We use the Foresee and Act procedure, which represents the social context with explicit mental states and observations using S$^{3}$AP. The agent proposes an action, predicts a possible next social state under that action, and uses the anticipated consequences to refine the proposal into the action it executes.
\end{enumerate}

\paragraph{Structured social representation.}

The Representor records the current state estimate in JSON as a set of social state hypotheses and their associated probability weights with natural language. In interactive tasks, it receives the complete history $H_t^\epsilon$ available before the current action and the ego's goal $G$, without receiving the current candidate actions. The resulting representation is shared across all candidate actions evaluated at that decision point.

The representation contains the following information:
\begin{itemize}
    \item World physical, at least includes physical conditions, state of objects, and events.
    \item Agent physical, at least includes location, involvement of events, and available information channels.
    \item Agent mental, at least include belief, goals, memories, relations, and estimate of others. Especially, this should include the agent observation.
\end{itemize}

Each representation should be either under evidence, which is the original text correspond to each fact or inference, or estimation. The state representation should not include field "uncertainty": facts relevant to the goal which is uncertainty for the agent's observation, including others' relevant mental states, should be represented as several possible states by the agent's possible inferences.

\paragraph{Interaction and state prediction.}

For each candidate action, the Social Predictor generates possible interactions through the process in Section~\ref{sec:predictor}. It receives structural represent and a candidate initial action by the ego agent. It first predicts the action's immediate physical consequences and then the participants' immediate mental updates conditioned on the physical outcome of the same branch. From this intermediate state, it predicts the other participants' responses and the resulting physical and mental successors with their possibilities. The output of Social Predictor is a set of possible future branches, each with a joint action of others and a future state with its possibility. A predicted respones may cause several possible future states, which are denoted as several future branches.

\paragraph{Action Planning.}

The Social Planner first proposes candidate actions directly from the ego's observable history, then evaluates $V(s_t,\hat s_{t+1},G)$ for each predicted successor state with each estimated current state. In each branch, this value represents the successor state's contribution to the ego's goal relative than the current state, considering both immediate progress and the conditions created or removed for future goal achievement. The Planner then refines the action according to branch values and transition probabilities of the candidate actions. The evaluation and action process is accomplished by a fixed LLM. The output of the Planner is one action.

\paragraph{State updates after interaction.}

After executing the action, the ego receives a new interaction record. An LLM then estimates its likelihood under the predicted interactions, as defined in Equation~\eqref{eq:likehood}. The Representor uses these likelihoods to update the state estimate in the next round through Bayesian according to Equation ~\eqref{eq:posterior-state-propagation}.

\paragraph{Judging criteria.}

Following the SOTOPIA evaluation protocol used by Foresee and Act (Zhou et al., 2026), we use a fixed LLM judge to evaluate each completed interaction. The judge receives the scenario context, the participants' character information and goals, and the interaction history. We use the same judge model, evaluation prompt, and scoring criteria across SocialDreamer, both baselines, and the ablation variants. The ego agent's score is on SOTOPIA dimensions, including goal achievement, believability, knowledge acquisition, relationship change, financial and material benefits, secret keeping, and adherence to social rules. On each dimension, higher score indicate better performance.

\subsection{Experimental Results}
\label{sec:experimental-results}

\paragraph{Overall performance.}

\begin{figure}[h]
\centering
\includegraphics[width=0.6\linewidth]
{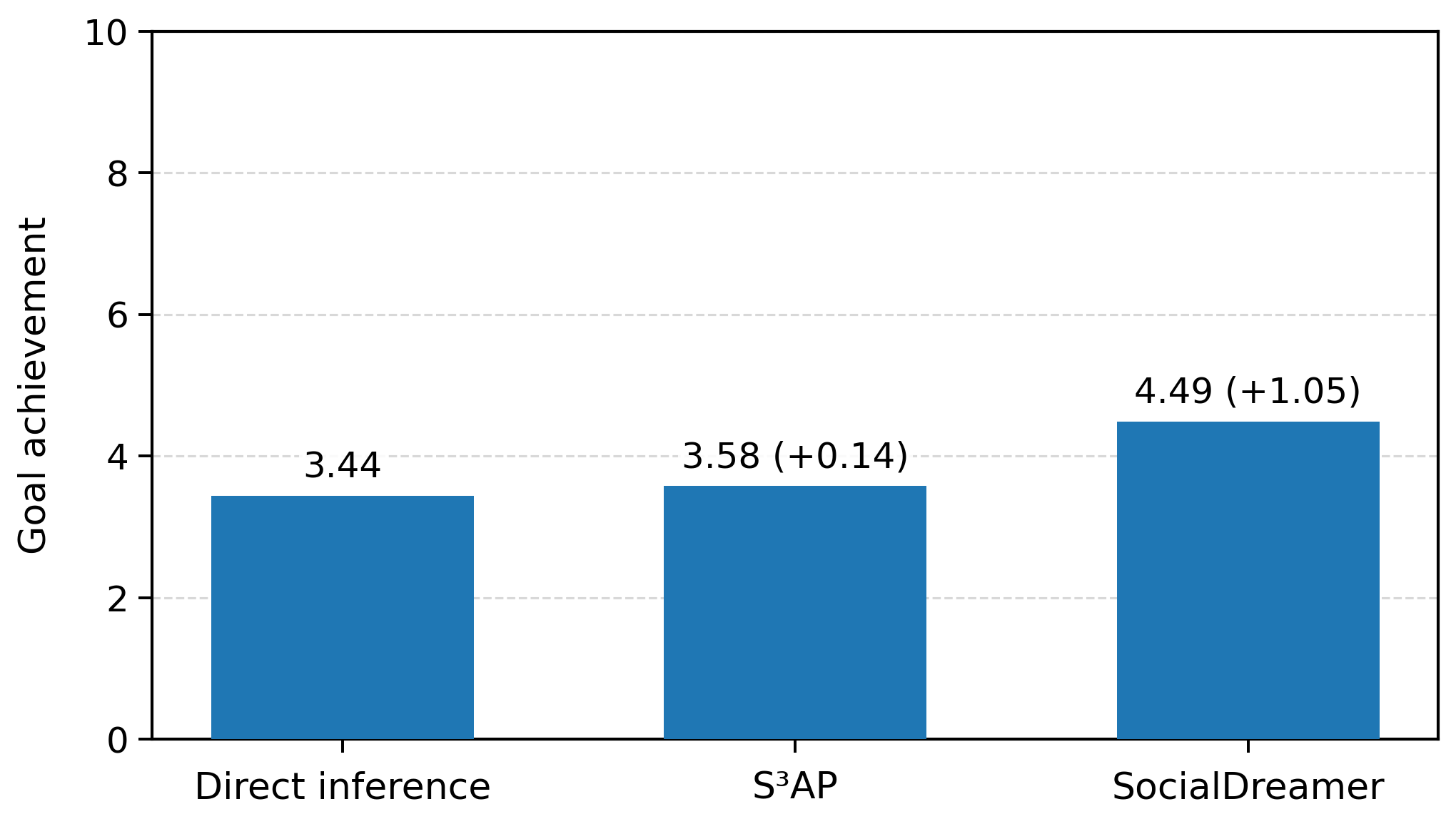}
\caption{Goal achievement on SOTOPIA-Hard.}
\label{fig:goal-results}
\end{figure}

We compare SocialDreamer with direct inference and S$^{3}$AP using the judging procedure described in Section~6.1. We use DeepSeek V4.1 Flash for SocialDreamer and both baselines, GPT-6.1 Terra as the interaction partner, and GPT-6.2 Sol as the judge. All models use fixed pretrained weights without additional post-training. Figure~\ref{fig:goal-results} reports the improvement in the ego agent's goal achievement score relative to each baseline. This comparison examines whether the SocialDreamer framework improves social decision-making over direct inference and the existing Foresee and Act procedure.

On SOTOPIA-Hard, SocialDreamer achieves a goal-achievement score of 4.49, representing a $30.5\%$ relative improvement over direct inference (3.44) and a $25.4\%$ relative improvement over Foresee and Act (3.58). Which is shown in Figure~\ref{fig:goal-results}.

Table~\ref{tab:main-results} further illustrate the score differences across all seven SOTOPIA-Hard evaluation dimensions. Including goal achievement, believability, knowledge acquisition, relationship change, financial and material benefits, secret keeping, and social rule adherence. These differences are consistent with the role of interaction prediction: anticipating how others may respond can help the ego identify actions that acquire useful information or create more favorable conditions for subsequent cooperation. Reporting all seven dimensions therefore provides a more complete picture of the consequences of social planning. Beyond goal achievement, SocialDreamer also has the highest average scores for knowledge acquisition, relationship change, and social rule adherence.

\begin{table}[h]
\centering
\small
\caption{Main comparison on SOTOPIA-Hard across all seven evaluation dimensions. Higher scores indicate better performance.}
\label{tab:main-results}
\begin{tabular}{lrrr}
\toprule
Evaluation Dimension & Direct & S$^{3}$AP & SocialDreamer \\
\midrule
Goal achievement (0--10)
& 3.44 & 3.58 & \textbf{4.49} \\

Believability (0--10)
& \textbf{8.49} & 7.86 & 7.72 \\

Knowledge acquisition (0--10)
& 5.24 & 5.40 & \textbf{5.78} \\

Relationship change (-5--5)
& 0.70 & 0.78 & \textbf{0.96} \\

Financial and material benefits (-5--5)
& \textbf{0.38} & 0.28 & -0.26 \\

Secret keeping (-10--0)
& -0.20 & -0.30 & -0.20 \\

Social rule adherence (-10--0)
& -0.32 & -0.33 & \textbf{-0.23} \\

\bottomrule
\end{tabular}
\end{table}

\begin{figure}[h]
\centering
\includegraphics[width=0.8\linewidth] {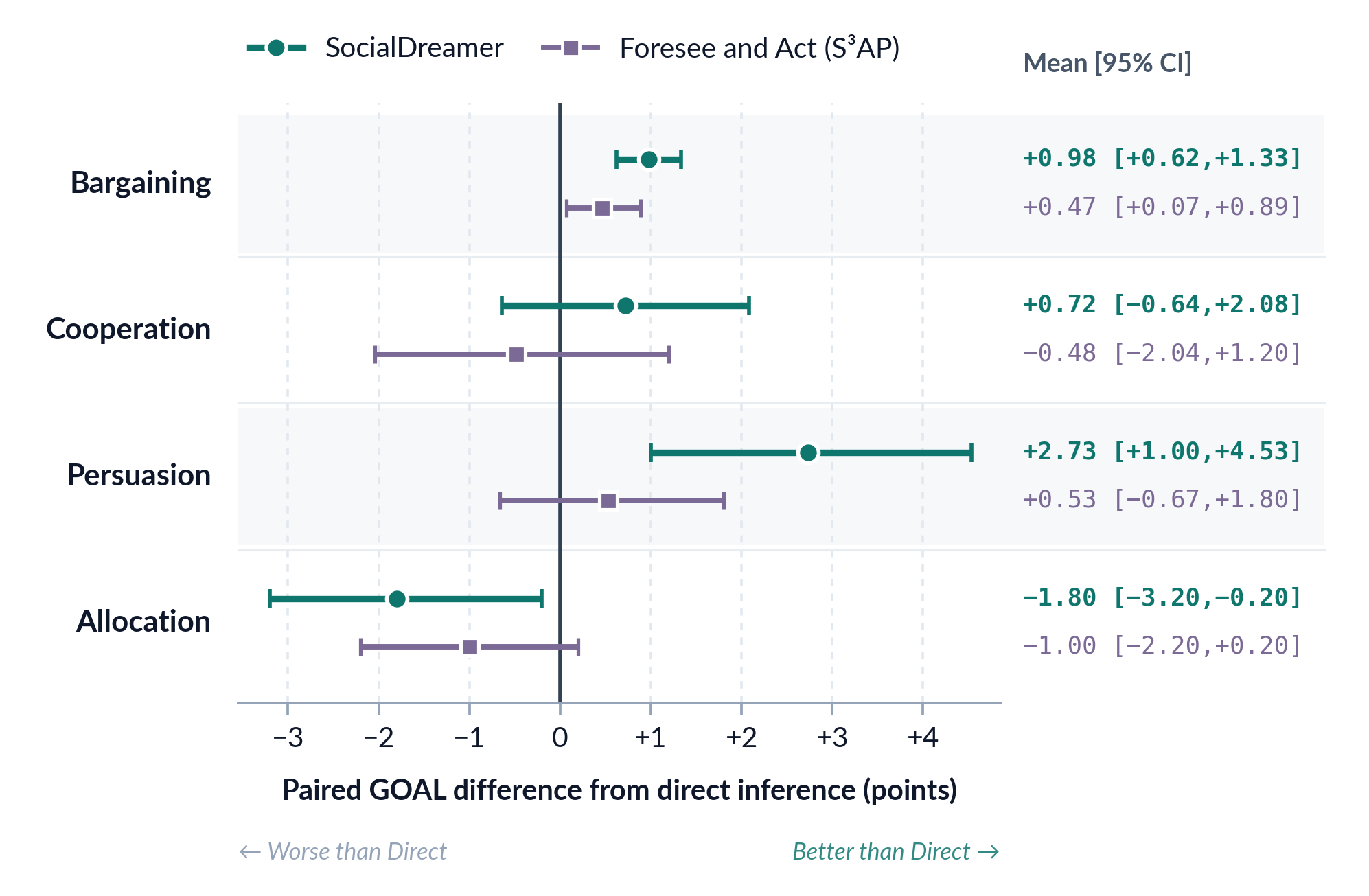}
\caption{\textbf{Paired goal-achievement differences from direct inference across SOTOPIA-Hard task types.} Points and bars show paired mean differences and 95\% case-bootstrap confidence intervals for SocialDreamer and Foresee and Act relative to direct inference. Positive values indicate higher goal achievement than direct inference.}
\label{fig:tasktypes}
\end{figure}

\paragraph{Goal achievement across task types.}
We grouped SOTOPIA-Hard into four task types by their primary goals: bargaining, cooperation, persuasion, and allocation. As shown in Figure~\ref{fig:tasktypes}, SocialDreamer achieves gains over the baselines across the three categories centered on communication and cooperation (bargaining, cooperation, and persuasion), whereas it underperforms in the competitive allocation setting.

On bargaining tasks, SocialDreamer is more willing to make concessions to reach agreements than S$^3$AP. Specifically, SocialDreamer closes a deal in $77.8\%$ of cases (35/45), more than doubling the $37.8\%$ completion rate (17/45) of S$^3$AP. However, its average goal-achievement score conditioned on completed deals is slightly lower than that of S$^3$AP (3.26 vs. 3.35). This pattern is consistent with a trade-off by prioritizing agreement completion over marginal price gains to avoid impasse. Thus, SocialDreamer scores higher on \textit{goal achievement} but lower on \textit{financial and material benefits}, compared to direct inference and S$^3$AP.

\paragraph{Ablation studies.}

We examine the contributions of Bayesian state updating and explicit interaction prediction. As Bayesian state updating depends on explicit interaction prediction, we use joint ablation for the latter. Table~\ref{tab:ablation-results} compares the full SocialDreamer with the two ablated variants and both baselines.
\begin{itemize} {
    \item\emph{Without Bayesian updating.} In this variant, the Representor reconstructs the state estimate from the complete interaction history at each round. This comparison examines whether explicitly linking observed feedback to previous predictions improves multi-round decision-making.

    \item\emph{Without explicit interaction prediction and Bayesian updating.} In this variant, the Social Predictor directly predicts successor states from the current-state hypotheses and candidate ego actions, without explicitly modeling the other participants' responses during action planning. This joint ablation further isolates the contribution of explicitly predicting interactions during planning.
}
\end{itemize}

\begin{figure}[h]
\centering
\includegraphics[width=0.95\linewidth]
{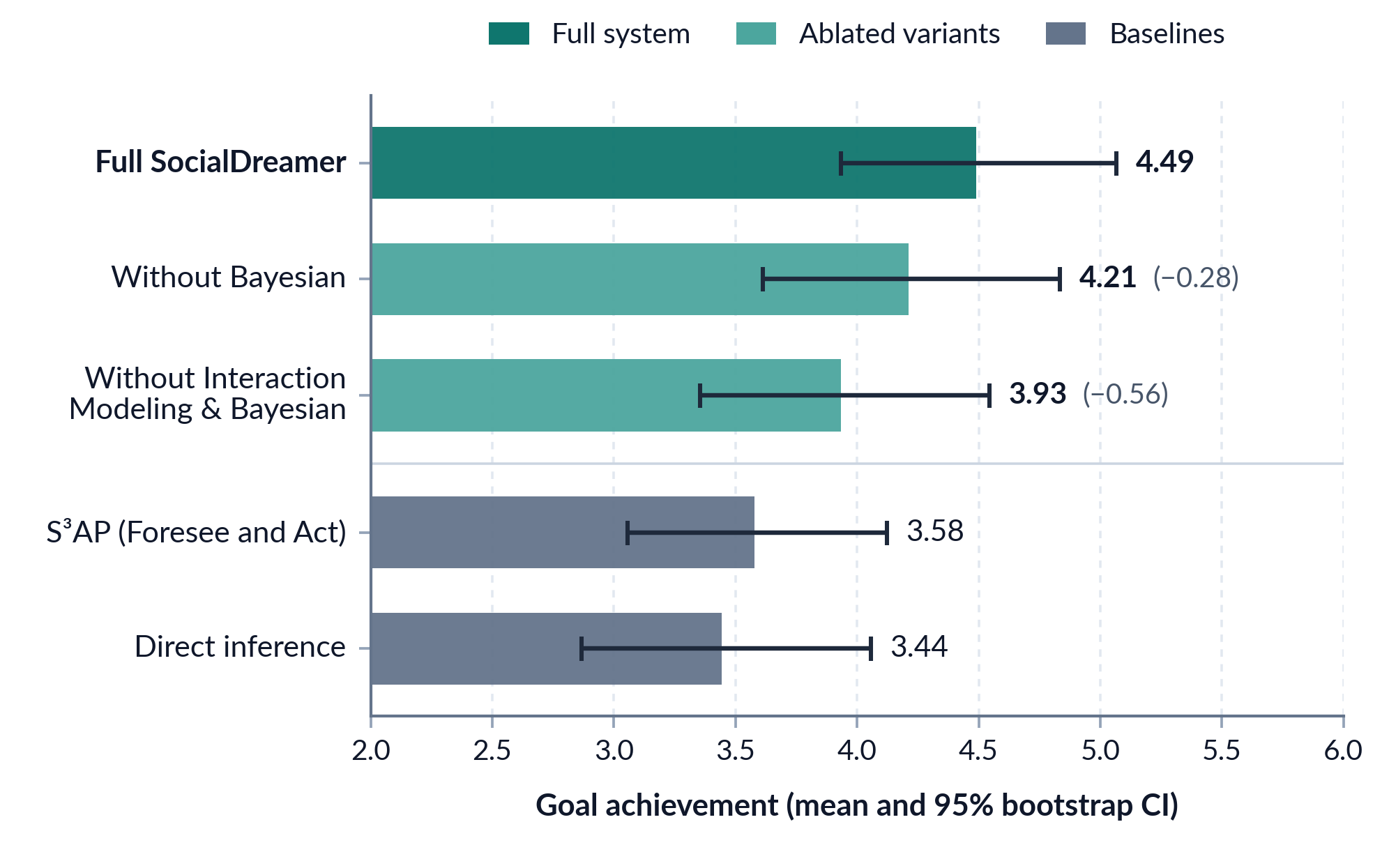}
\caption{Ablation analysis on SOTOPIA-Hard showing mean GOAL achievement scores (out of 10) for the full SocialDreamer, its two ablated variants, and the two baselines. Note that the X-axis is magnified to $[2.0, 6.0]$ interval to highlight performance differences. Error bars show 95\% confidence intervals.}
\label{fig:mechanism-ablation}
\end{figure}

As shown in Figure~\ref{fig:mechanism-ablation}, the ablation results confirm that both components contribute to goal achievement. Removing Bayesian state updating reduces the mean GOAL score from 4.49 to 4.21, and further removing explicit interaction modeling lowers it to 3.93. Even with both components removed, the ablated variant still outperforms S$^3$AP (3.58) and direct inference (3.44). These results illustrate the complementary contributions of explicit interaction prediction and Bayesian state updating. The former allows the Planner to anticipate how other participants may respond before committing to an action, while the latter uses actual responses to revise the ego's estimates of the social world across rounds.

Across all seven dimensions in Table~\ref{tab:ablation-results}, both components also benefit knowledge acquisition (KNO: 5.57 to 5.78) and relationship quality (REL: 0.62 to 0.96). Specifically, Bayesian state updating reflects a goal-driven trade-off: removing it raises financial and material benefits (FIN: $-0.26$ to $0.06$) at the expense of goal achievement (4.49 to 4.21) and relationship scores (0.96 to 0.86). Further removing explicit interaction modeling causes the sharpest drops in relationship (0.86 to 0.62) and social norm compliance (SOC: $-0.23$ to $-0.49$). Without anticipating partner responses before acting, the agent is less able to preserve rapport and social boundaries across turns.

\begin{table}[H]
\centering
\small
\caption{Ablation results on SOTOPIA-Hard.}
\label{tab:ablation-results}
\begin{tabular}{lrrrrrrr}
\toprule
Method & GOAL & BEL & KNO & REL & FIN & SEC & SOC \\
\midrule
Direct inference
& 3.44 & 8.49 & 5.24 & 0.70 & 0.38 & -0.20 & -0.32 \\

S$^3$AP
& 3.58 & 7.86 & 5.40 & 0.78 & 0.28 & -0.30 & -0.33 \\

SocialDreamer
& \textbf{4.49} & 7.72 & \textbf{5.78}
& \textbf{0.96} & -0.26 & -0.20 & -0.23 \\

\midrule
Without Bayesian.
& 4.21 & 7.58 & 5.70 & 0.86 & 0.06 & -0.12 & -0.23 \\

\midrule
\begin{tabular}[c]{@{}l@{}}Without Interaction \\ Modeling \& Bayesian\end{tabular}
& 3.93 & 7.59 & 5.57 & 0.62 & 0.04 & -0.26 & -0.49 \\
\bottomrule
\end{tabular}
\end{table}

\section{Conclusion}
\label{sec:conclusion}

SocialDreamer extends Dreamer's idea of guiding behavior through imagined futures to social settings. Recognizing that social dynamics depend on hidden mental states and distinct information boundaries, SocialDreamer organizes the global social world into a structured state representation. Upon this representation, the Social Predictor models one-round transition predictions with explicitly predicting others' responses. The Social Representor maintains a distribution over structured state hypotheses from the ego's interaction history, and recursively updates it via Bayesian inferenced by predicted and newly observed interactions. Finally, the Social Planner constructs candidate actions, evaluates imagined futures against its goal, and refines the action.

To evaluate the structural design of SocialDreamer without confounding effects from model post-training, we instantiated the framework using fixed language models with explicit structured states. On SOTOPIA-Hard, SocialDreamer achieves a mean goal-achievement score of 4.49, compared with 3.44 for direct inference and 3.58 for Foresee and Act.

The key direction for future work is to learn the Social Representor, Social Predictor, and Social Planner jointly on interaction data. This may further improve predictive accuracy, probability calibration, and inference efficiency, whether through fine-tuning language models or learning compact latent dynamics. Extending SocialDreamer from one-round prediction to multi-round lookahead is another important direction. Simulating longer imagined trajectories alongside a learned continuation value estimator would allow the ego agent to anticipate longer-term social consequences and make farsighted decisions.

\section*{References}

\noindent\hangindent=2em
Baker, C. L., Saxe, R., \& Tenenbaum, J. B. (2009). Action understanding as inverse planning. \textit{Cognition, 113}(3), 329--349. \url{https://doi.org/10.1016/j.cognition.2009.07.005}

\vspace{0.5em}
\noindent\hangindent=2em
Doucet, A., Godsill, S., \& Andrieu, C. (2000). On sequential Monte Carlo sampling methods for Bayesian filtering. \textit{Statistics and Computing, 10}, 197--208. \url{https://doi.org/10.1023/A:1008935410038}

\vspace{0.5em}
\noindent\hangindent=2em
Fei, H., \& Zhao, Y. (2026). Mental world modeling. \textit{arXiv preprint arXiv:2607.27201}. \url{https://doi.org/10.48550/arXiv.2607.27201}

\vspace{0.5em}
\noindent\hangindent=2em
Givan, R., Dean, T., \& Greig, M. (2003). Equivalence notions and model minimization in Markov decision processes. \textit{Artificial Intelligence, 147}(1--2), 163--223. \url{https://doi.org/10.1016/S0004-3702(02)00376-4}

\vspace{0.5em}
\noindent\hangindent=2em
Gmytrasiewicz, P. J., \& Doshi, P. (2005). A framework for sequential planning in multi-agent settings. \textit{Journal of Artificial Intelligence Research, 24}, 49--79. \url{https://doi.org/10.1613/jair.1579}

\vspace{0.5em}
\noindent\hangindent=2em
Hafner, D., Lillicrap, T., Ba, J., \& Norouzi, M. (2020). Dream to control: Learning behaviors by latent imagination. In \textit{International Conference on Learning Representations (ICLR)}. \url{https://doi.org/10.48550/arXiv.1912.01603}

\vspace{0.5em}
\noindent\hangindent=2em
Hafner, D., Pasukonis, J., Ba, J., \& Lillicrap, T. (2025). Mastering diverse control tasks through world models. \textit{Nature, 640}, 647--653. \url{https://doi.org/10.1038/s41586-025-08744-2}

\vspace{0.5em}
\noindent\hangindent=2em
Kaelbling, L. P., Littman, M. L., \& Cassandra, A. R. (1998). Planning and acting in partially observable stochastic domains. \textit{Artificial Intelligence, 101}(1--2), 99--134. \url{https://doi.org/10.1016/S0004-3702(98)00023-X}

\vspace{0.5em}
\noindent\hangindent=2em
Premack, D., \& Woodruff, G. (1978). Does the chimpanzee have a theory of mind? \textit{Behavioral and Brain Sciences, 1}(4), 515--526. \url{https://doi.org/10.1017/S0140525X00076512}

\vspace{0.5em}
\noindent\hangindent=2em
Rabinowitz, N., Perbet, F., Song, F., Zhang, C., Eslami, S. M. A., \& Botvinick, M. (2018). Machine theory of mind. In \textit{Proceedings of the 35th International Conference on Machine Learning} (Vol. 80, pp. 4218--4227). PMLR. \url{https://doi.org/10.48550/arXiv.1802.07740}

\vspace{0.5em}
\noindent\hangindent=2em
Ross, S., Pineau, J., Paquet, S., \& Chaib-draa, B. (2008). Online planning algorithms for POMDPs. \textit{Journal of Artificial Intelligence Research, 32}, 663--704. \url{https://doi.org/10.1613/jair.2567}

\vspace{0.5em}
\noindent\hangindent=2em
Zheng, L., Chiang, W.-L., Sheng, Y., Zhuang, S., Wu, Z., Zhuang, Y., Lin, Z., Li, Z., Li, D., Xing, E. P., Zhang, H., Gonzalez, J. E., \& Stoica, I. (2023). Judging LLM-as-a-judge with MT-Bench and Chatbot Arena. In \textit{Advances in Neural Information Processing Systems} (Vol. 36). \url{https://doi.org/10.52202/075280-2020}

\vspace{0.5em}
\noindent\hangindent=2em
Zhou, X., Liu, J., Yerukola, A., Kim, H., \& Sap, M. (2026). Social world models. In \textit{Conference on Language Modeling (COLM)}. \url{https://doi.org/10.48550/arXiv.2509.00559}

\vspace{0.5em}
\noindent\hangindent=2em
Zhou, X., Zhu, H., Mathur, L., Zhang, R., Yu, H., Qi, Z., Morency, L.-P., Bisk, Y., Fried, D., Neubig, G., \& Sap, M. (2024). SOTOPIA: Interactive evaluation for social intelligence in language agents. In \textit{International Conference on Learning Representations (ICLR)}. \url{https://doi.org/10.48550/arXiv.2310.11667}

\newpage

\begin{appendices}
\section{Immediate and Long-Term Value}
\label{app:value}

The value introduced in Section \ref{sec: evaluation} combines the immediate consequences of an action with the opportunities it creates for subsequent decisions. Let $r(s_t,\tilde s_{t+1},G)$ denote the immediate goal-conditioned reward associated with one interaction round, while long-term consequences are evaluated through subsequent decisions based on the ego's updated information.

With a discount factor $\gamma\in[0,1)$, the transition value is decomposed as
\begin{equation}
\begin{aligned}
V(s_t,\tilde s_{t+1},G)
={}&r(s_t,\tilde s_{t+1},G)
\\
&+\gamma
V\left(
a_{t+1}^{\epsilon,chosen},
\hat H_{t+1}^\epsilon,
G
\right).
\end{aligned}
\label{eq:immediate-continuation-value}
\end{equation}
In which $\hat H_{t+1}^\epsilon$ contains the information that would become available to the ego in the predicted interaction branch. This describes the value within an individual predicted interaction branch. When multiple interaction records correspond to the same state transition, the transition value used in Equation~\eqref{eq:value} averages over their corresponding continuation values.

To evaluate multi-rounds interaction explicitly, the Planner generates imagined trajectories of length $L$. Each trajectory starts from a possible current state $s_t\sim R_\phi(\cdot\mid H_t^\epsilon)$, with the initiating action fixed to the candidate $a_t^\epsilon$ being evaluated. At each imagined round $t+h$, for $h=0,\ldots,L-1$, the Predictor generates the other participants' responses and the successor
\[
\left(\hat a_{t+h}^{-\epsilon},\hat s_{t+h+1}\right)\sim K_\theta\left(\cdot,\cdot \mid \tilde s_{t+h},a_{t+h}^\epsilon \right).
\]
We initialize $\tilde s_t=s_t$ and $\hat H_t^\epsilon=H_t^\epsilon$. For the rollout described here, we adopt the simplification in Section~\ref{sec:planning-bounds} and set $\tilde s_{t+h+1}=\hat s_{t+h+1}$ at each imagined round.

The information available to the ego within that branch extends its imagined history: 
\[
\hat H_{t+h+1}^\epsilon = \hat H_{t+h}^\epsilon\oplus\Delta\hat H_{t+h+1}^\epsilon.
\]

The multi-round value is approximated by
\begin{equation}
V(a_t^\epsilon,H_t^\epsilon,G) \approx \mathbb E\Bigg[ \sum_{h=0}^{L-1}
\gamma^h r\left( \tilde s_{t+h}, \tilde s_{t+h+1}, G \right)
+\gamma^L V\left(a_{t+L}^{\epsilon,chosen},\hat H_{t+L}^\epsilon,G\right)].
\label{eq:multi-round-value}
\end{equation}
The expectation is approximated by averaging returns over multiple imagined trajectories. All trajectories for a candidate share the same initial ego action, while later actions may differ in response to different ego-visible evidence. Increasing $L$ extends explicit lookahead, while the terminal value allows shorter rollouts to account for consequences beyond their simulated horizon.

The immediate reward and continuation value may be specified by the task or learned from returns constructed using imagined rewards and subsequent value estimates. After planning, only the selected current-round action is executed. The actual interaction then updates $H_{t+1}^\epsilon$ and the ego's state estimate, and a new planning cycle begins.

\section{Confidence and Support Score}
\label{app:confidence}

Let $c(s_t,\hat s_{t+1})\in[0,1]$ denote the support score for a predicted branch and its proposed representative. With a specified common lower bound $V_{\min}$ for the evaluated branch values given by the task, we define the adjusted value:
\begin{equation}
\begin{aligned}
V^c(a_t^\epsilon,H_t^\epsilon,&G) = 
\int_{\mathcal S} R_\phi(s_t\mid H_t^\epsilon) \times 
 \Bigg[\int_{\mathcal S} P_\theta(\hat s_{t+1}\mid s_t,a_t^\epsilon) \\ & \times \left(c(s_t,\hat s_{t+1}) \times V(s_t,\tilde s_{t+1},G)+(1-c(s_t,\hat s_{t+1}))V_{\min}\right)\mathrm d\hat s_{t+1}\Bigg]
\mathrm ds_t.
\end{aligned}
\label{eq:confidence-constrained-value}
\end{equation}
Here, representative $\tilde s_{t+1}$ belongs to the equivalence class of its predicted successor $\hat s_{t+1}$ and satisfies the constraints of behavioral support. The credibility score $c$ is evaluated for the original predicted interaction branch.

For any branch value satisfying $V(s_t,\tilde s_{t+1},G)\geq V_{\min}$, the adjusted score satisfies:
\[
V_{\min} \leq cV(s_t,\tilde s_{t+1},G)+(1-c)V_{\min} \leq V(s_t,\tilde s_{t+1},G).
\]
Thus, predictions with weaker credibility contribute more conservatively to the expected value, rather than being discarded from the predictive distribution.

In addition, we require the average credibility of a candidate action to exceed a predefined threshold $c_{\min}$. The average credibility is

\begin{equation}
\bar c(a_t^\epsilon) ={}\int_{\mathcal S}R_\phi(s_t\mid H_t^\epsilon)\Bigg[\int_{\mathcal S}P_\theta\left(\hat s_{t+1}\mid s_t,a_t^\epsilon\right) \times c(s_t,\hat s_{t+1}) \mathrm d\hat s_{t+1}\Bigg]\mathrm ds_t.
\label{eq:average-credibility}
\end{equation}

We may evaluate the expectations using Monte Carlo samples. For each candidate action $a_t^\epsilon$, generate $W \times M$ predictive branches \(s_t^{(w)} \sim R_\phi(\cdot\mid H_t^\epsilon)\) and \((\hat s_{t+1}^{(w,m)}) \sim P_\theta \left(
\cdot \mid s_t^{(w)},a_t^\epsilon\right)\) for each $s_t^{(w)}$. Let $c^{(w,m)}$ denote the credibility score of the $(w,m)$-th predicted branch. The confidence-adjusted value is approximated by
\begin{equation}
V^c(a_t^\epsilon,H_t^\epsilon,G)
\approx \frac{1}{WM}\sum_{w=1}^{W}\sum_{m=1}^{M}
\left[c^{(w,m)}V\left(s_t^{(w)},\tilde s_{t+1}^{(w,m)},G\right)
+(1-c^{(w,m)})V_{\min}\right].
\label{eq:mc-confidence-value}
\end{equation}

\end{appendices}

\end{document}